\def\year{2022}\relax
\documentclass[letterpaper]{article} 
\usepackage{aaai22}  
\usepackage{times}  
\usepackage{helvet}  
\usepackage{courier}  
\usepackage[hyphens]{url}  
\usepackage{graphicx} 
\usepackage{natbib}  
\usepackage{caption} 
\DeclareCaptionStyle{ruled}{labelfont=normalfont,labelsep=colon,strut=off} 
\usepackage{algorithm}
\usepackage{algorithmic}
\usepackage{multirow}
\usepackage{multicol}
\usepackage{amssymb}
\usepackage{amsmath}
\usepackage{color}
\usepackage{appendix}
\usepackage{stfloats}
\usepackage{footmisc}
\usepackage{newfloat}
\usepackage{listings}
\floatstyle{ruled}
\newfloat{listing}{tb}{lst}{}
\floatname{listing}{Listing}

\usepackage[switch]{lineno}
\usepackage{lipsum}
\usepackage{hyperref} 
\usepackage{xcolor}

\title{Knowledge Distillation for Object Detection via Rank Mimicking and Prediction-guided Feature Imitation}
\author {
    Gang Li\textsuperscript{\rm 1,2}\footnote{Equal contribution.},
    Xiang Li\textsuperscript{\rm 1}\footnotemark[1],
    Yujie Wang\textsuperscript{\rm 2},
    Shanshan Zhang\textsuperscript{\rm 1}\footnote{Corresponding author.},
    Yichao Wu\textsuperscript{\rm 2},
    Ding Liang\textsuperscript{\rm 2}
}

\affiliations {
    \textsuperscript{\rm 1} Nanjing University of Science and Technology \\
    \textsuperscript{\rm 2} Sensetime Research \\
    \{gang.li,~xiang.li.implus,~shanshan.zhang\}@njust.edu.cn, \{wangyujie,~wuyichao,~liangding\}@sensetime.com
}

\usepackage{bibentry}

\begin{document}
\maketitle
\begin{abstract}

Knowledge Distillation (KD) is a widely-used technology to inherit information from cumbersome teacher models to compact student models, consequently realizing model compression and acceleration.
Compared with image classification, object detection is a more complex task, and designing specific KD methods for object detection is non-trivial.   
In this work, we elaborately study the behaviour difference between the teacher and student detection models, and obtain two intriguing observations:
First, the teacher and student rank their detected candidate boxes quite differently, which results in their precision discrepancy.
Second, there is a considerable gap between the feature response differences and prediction differences between teacher and student, indicating that equally imitating all the feature maps of the teacher is the sub-optimal choice for improving the student's accuracy.
Based on the two observations, we propose Rank Mimicking (RM) and Prediction-guided Feature Imitation (PFI) for distilling one-stage detectors, respectively. 
RM takes the rank of candidate boxes from teachers as a new form of knowledge to distill, which consistently outperforms the traditional soft label distillation.  
PFI attempts to correlate feature differences with prediction differences, making feature imitation directly help to improve the student's accuracy. 
On MS COCO and PASCAL VOC benchmarks, extensive experiments are conducted on various detectors with different backbones to validate the effectiveness of our method. Specifically, RetinaNet with ResNet50 achieves 40.4\% mAP in MS COCO, which is 3.5\% higher than its baseline, and also outperforms previous KD methods. 

\end{abstract}

\section{Introduction}

As a fundamental computer vision task, object detection attracts much attention and achieves significant progress in recent years~\cite{gfl,varifocal}. It also has extensive applications in the real world, such as autonomous driving, intelligent surveillance and robotics. With the development of deep learning, Convolutional Neural Networks (CNNs) based detectors dominate the field and replace the traditional hand-craft detectors. When equipped with high-capacity CNN backbones, the advanced detectors can obtain satisfactory performance, however, these cumbersome models hardly satisfy real-time requirements and are useless in practical deployments. 
To address this dilemma, some researchers attempt to design efficient detectors, such as YOLO~\cite{yolo} and EfficientDet~\cite{efficientdet}. 
Apart from this, another solution is to transfer knowledge from cumbersome teacher models to lightweight student models, and help students to achieve comparable performance with teachers, which is known as Knowledge Distillation (KD)~\cite{hinton}. We focus on KD in this paper. 

\begin{figure}
    \centering
    \includegraphics[width=0.46\textwidth]{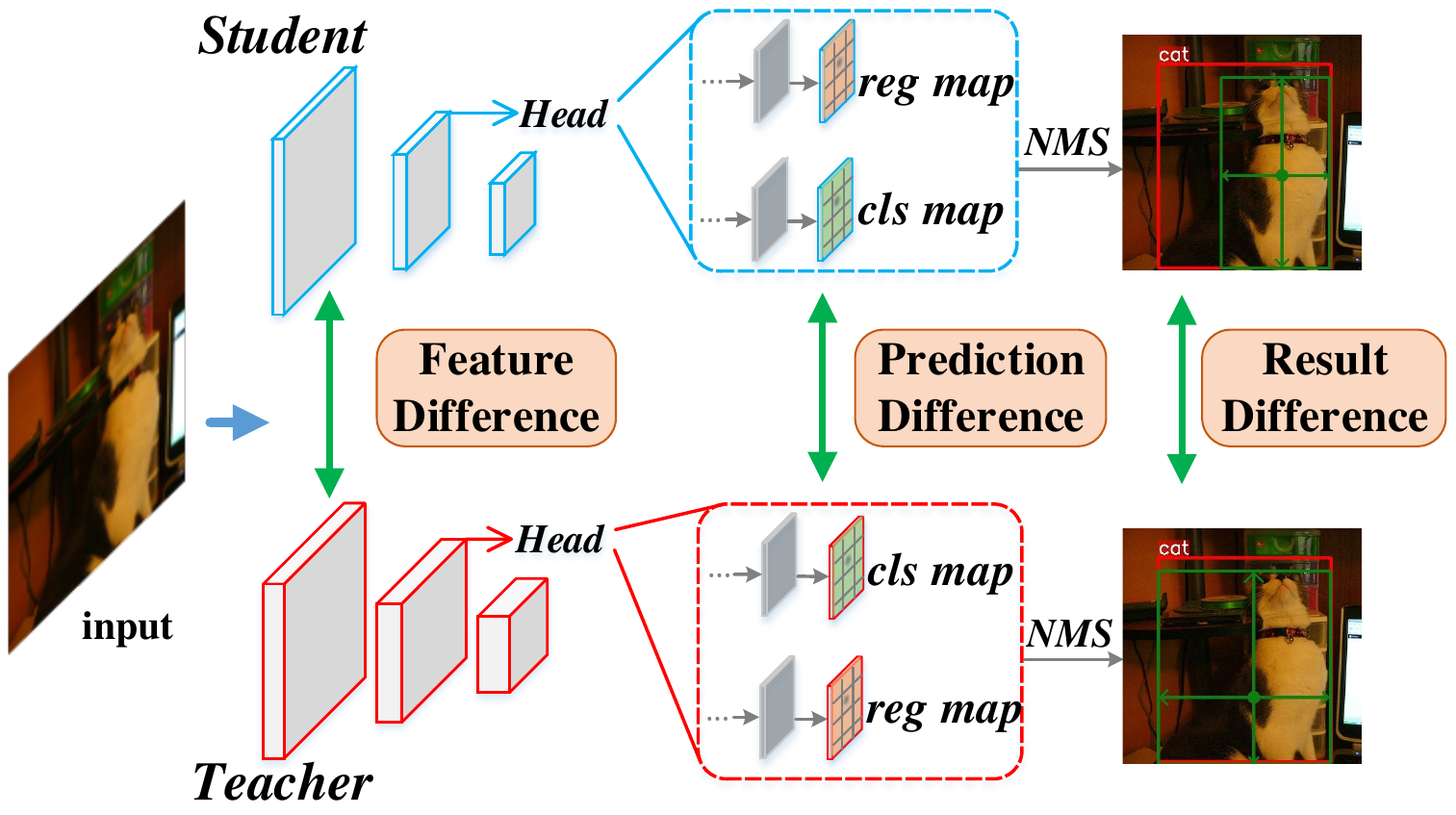}
    \caption{To figure out where the performance gap between teachers and students comes from, in this paper, we elaborately compare their behaviour, including the feature difference, prediction difference and detection result difference.}
    \label{fig:structure}
\end{figure}

\begin{figure}
    \centering
    \includegraphics[width=0.46\textwidth]{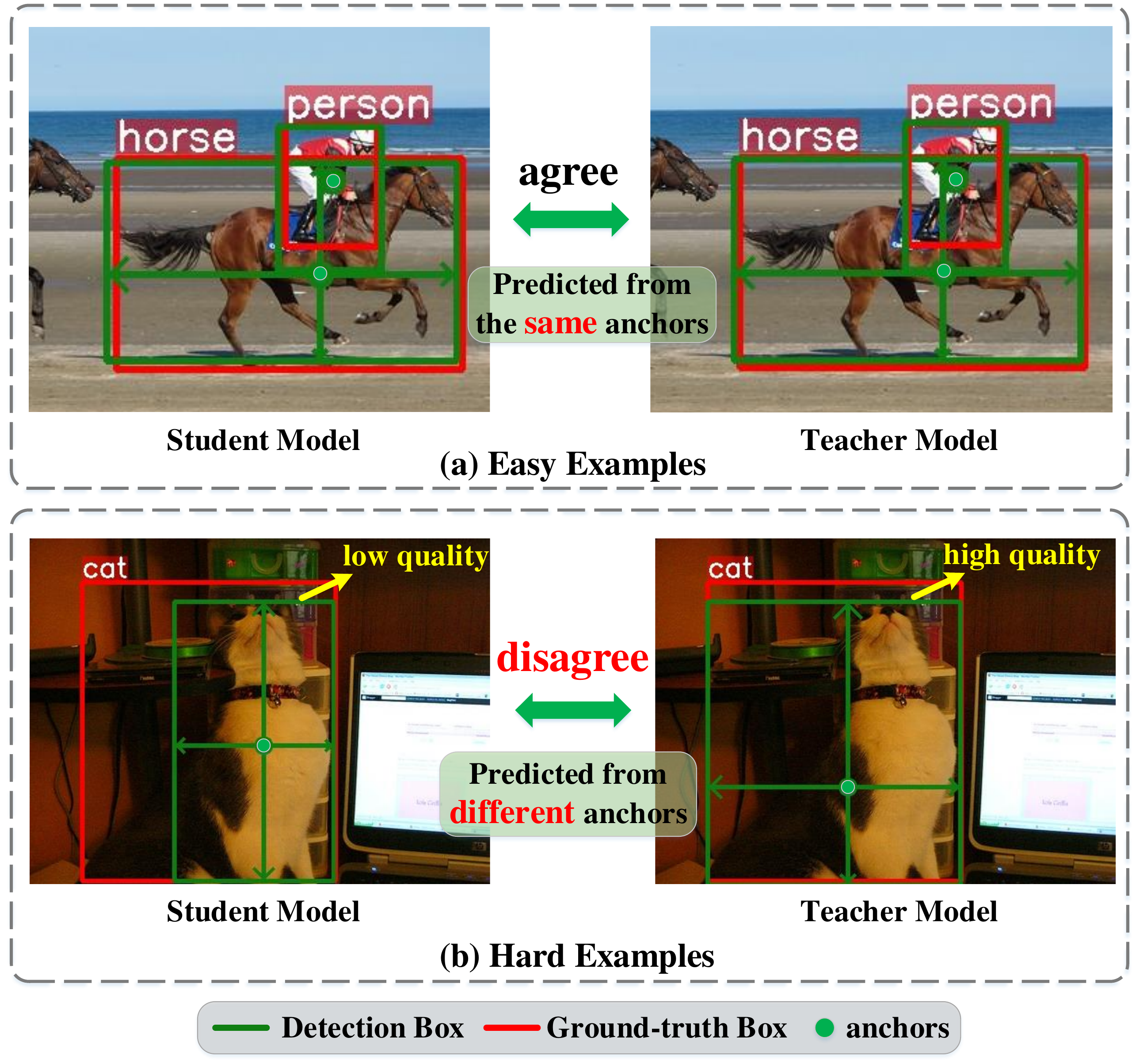}
    \caption{Comparison between the detection results from teachers and students. The final detection boxes (green bounding boxes) preserved from NMS and its corresponding anchors are plotted. (a) For easy examples, both teachers and students predict precise boxes. (b) For hard examples, students fail to generate high-quality boxes.}
    \label{fig:motivation_RM}
\end{figure}

KD is firstly proposed to compress classification networks, but in the past few years, researchers adapt it to object detection. Mimicking~\cite{mimicking} only adopts proposal features from teachers as distillation targets; Chen et al.~\cite{efficientKD} borrow the idea of soft labels from ~\cite{hinton}, and directly apply it to the classification branch of detection; ~\cite{FGFI,TAR} design fine-grained masks for feature imitation; DeFeat~\cite{deFeat} investigates the balance of foreground and background in the process of distillation; and GID~\cite{GID} proposes to perform KD in an instance-wise manner. 
Although these works have made encouraging progress on detection knowledge distillation, they fail to conduct a deep investigation on the behaviour difference between teacher and student.

 


In this paper, we elaborately compare the behaviour of the teacher and student detection models, including the final detection results (after NMS), dense predictions and feature responses, as shown in Fig.~\ref{fig:structure}. Through these comparisons, we obtain the following two interesting observations:

\textbf{(1) The teacher and student rank their candidate detections quite differently.}
For teacher and student detection models, we compare their final detection boxes along with their corresponding anchors\footnote{Anchor points for an anchor-free detector, or anchor boxes for an anchor-based detector} in Fig.~\ref{fig:motivation_RM}: 
for easy examples, both teachers and students can make perfect predictions and \emph{the final detection boxes are produced from the same anchors}; whilst for hard examples, disagreements between students and teachers appear, i.e., \emph{the final detection box of the student is inferior and regressed from a distinct anchor}. This phenomenon indicates that for hard examples, students have different anchor ranks from teachers, which may result in their performance gap. According to the observation, we propose Rank Mimicking (RM), which can transfer a new form of knowledge, i.e., the rank distribution of anchors, from teacher to student.  

\begin{figure*}
    \centering
    \includegraphics[width=0.9\textwidth]{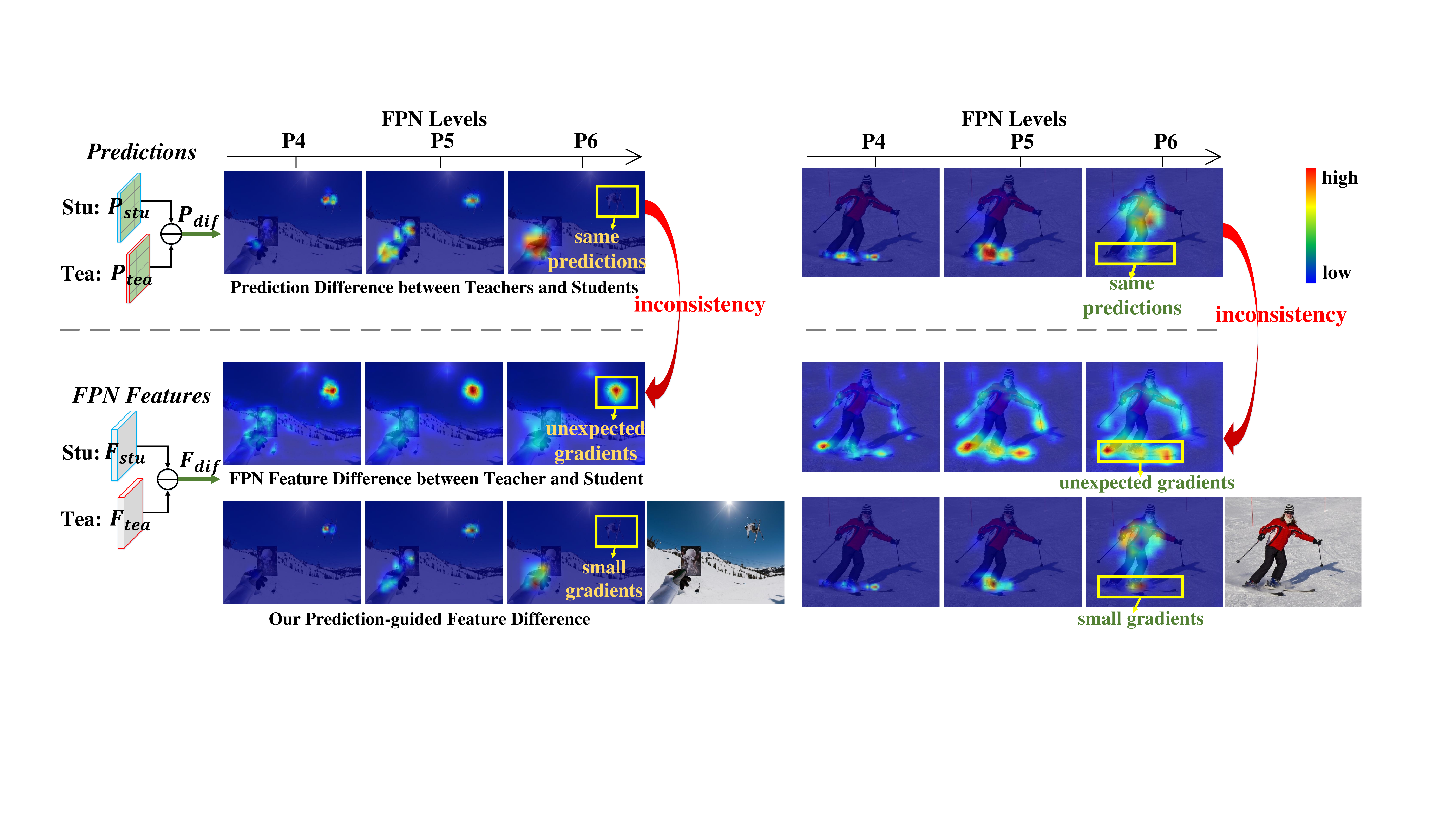}
    \caption{Two examples to illustrate the vanilla feature imitation does not directly contribute to accurate prediction.
    The 1st row: the prediction difference between teacher and student; 
    the 2nd row: the FPN feature difference; 
    the 3rd row: the calibrated feature difference by our PFI.
    The feature difference is positively correlated with distillation gradients.}
    \label{fig:motivation_PFI}
\end{figure*}

\textbf{(2) A notable gap between the feature differences and prediction differences between teacher and student.}
In Fig.~\ref{fig:motivation_PFI}, we visualize the prediction differences ($P_{dif} = P_{tea} - P_{stu}$) and feature differences ($F_{dif} = F_{tea} - F_{stu}$) between teachers and students. 
From the first two rows of Fig.~\ref{fig:motivation_PFI}, it is obvious that the feature differences $F_{dif}$ are apparently inconsistent with the prediction ones $P_{dif}$. Specifically, inside the yellow boxes, students can already generate as precise predictions as teachers ($P_{dif}$ is small), but the corresponding $F_{dif}$ (typically $F_{dif}$ is positively related to KD supervisions across feature maps) is large, which may lead to unexpected or wasted gradients during feature imitation. 
To make feature imitation directly contribute to accurate prediction for students, we propose Prediction-guided Feature Imitation, where we incorporate $P_{dif}$ as position-wise loss weights to guide the efficient and accurate feature imitation.    

To validate the effectiveness of our method, we conduct extensive experiments on two widely-used object detection datasets, MS COCO~\cite{coco} and PASCAL VOC~\cite{pascal}. 
Through the proposed KD methods, ResNet50 based RetinaNet~\cite{focal} achieves 40.4\% AP, which surpasses the baseline by 3.5\%, and it also outperforms the previous state-of-the-art methods by a large margin. 

To summarize, our contributions are as follows:
\begin{itemize}
    \item Through elaborately comparing the behaviour of teacher and student detection models, we obtain two meaningful observations, which have the potential to guide and inspire further designs on detection knowledge distillation. 
    \item A novel effective type of knowledge, i.e., rank distribution of anchors is introduced, and it consistently outperforms the traditional soft label distillation through the proposed Rank Mimicking (RM) process. 
    \item The proposed Prediction-guided Feature Imitation (PFI) is the first to employ the prediction difference to guide efficient feature imitation, which closely correlates feature imitation with accurate predictions.
    \item RM and PFI can significantly improve object detection, and outperform other distillation methods on MS COCO and PASCAL VOC benchmarks. With ResNet152 based RetinaNet as the teacher (40.1\% AP), our ResNet50 based student improves the baseline from 36.9\% to 40.4\% on COCO. 
\end{itemize}

\section{Related Work}

\subsection{Object Detection}

Object detection frameworks can be roughly divided into two categories: one-stage and two-stage detectors. In the early years, two-stage detectors dominate the detection field due to superior performance. Faster R-CNN~\cite{fasterRCNN} designs Region Proposal Network to generate candidate boxes in a fully convolutional manner, then uses RoI Pooling to crop features for sampled proposals and performs prediction refinement on these features. Cascade R-CNN~\cite{cascadeRCNN} proposes multiple sequential R-CNN heads to perform multiple refinements on proposals, so as to obtain more precise detection boxes. 
However, in the last two years, one-stage detectors become popular thanks to their compact structure and low latency. Lin et al. propose Focal Loss~\cite{focal} to address the severe imbalance problem between positive and negative samples, and significantly narrows the performance gap between two-stage and one-stage detectors. FCOS~\cite{FCOS} designs an anchor-free framework, removing the pre-defined anchors and directly predicting boxes from feature map points, which contributes to a clear and simple detection framework. ATSS~\cite{atss} proposes the label assignment is the intrinsic difference between anchor-free and anchor-based detectors, and it introduces adaptive training sample selection to assign positive anchors to instances in a dynamic manner. GFL~\cite{gfl} introduces quality focal loss (QFL) and distribution focal loss (DFL), where QFL replaces the target label with the quality indicator (IoU), leading to a more accurate box rank, and DFL estimates bounding boxes by modeling regression offsets as general distributions. Considering that one-stage detectors are more efficient, practical and easier to deploy, we focus on knowledge distillation for one-stage detectors in this work. 

\subsection{Knowledge Distillation for Object Detection}

Knowledge Distillation (KD) is formally proposed by Hinton et al.~\cite{hinton} to pass dark knowledge from complicated teachers to compact students, enabling students to maintain strong performance as teachers. Besides image classification, KD is also widely applied to object detection tasks recently. Mimicking~\cite{mimicking} firstly extends KD methods to object detection, specifically, it attempts to employ traditional logits mimic learning to object detection, but fails and gets worse performance than supervised by GT only. Then it proposes to mimic features samples from regions of proposals. ~\cite{efficientKD} views the intermediate feature imitation as hint learning, and designs a feature adaptation layer to align the dimension of the hint layer and guided layer, which also narrows the semantic gap between the hint and guided layers.
FGFI~\cite{FGFI} claims detectors care more about local regions that overlap with ground truth objects, thus it estimates the imitation mask to only imitate features from the regions near object anchor locations. 
deFeat~\cite{deFeat} points out the features derived from regions excluding objects also carry import information to distill students and can suppress false positives from the background. It carefully assigns different loss weights to foreground and background regions during the imitation. 
GID~\cite{GID} proposes to distill student detectors in an instance-wise manner. It firstly generates proposals for instances, then uses RoI Align to crop instance features. At last, feature-based, response-based and relation-based distillation are performed on instance features.

In summary, previous feature imitation methods heavily rely on hand-craft region selection, specifically, they either adopt proposal regions~\cite{mimicking,GID}, or carefully select foreground regions based on priors~\cite{FGFI}. Besides, for output distillation, they all follow traditional soft label distillation~\cite{hinton}. 
Therefore, compared with these methods, we summarize our main differences as follows:
(1) Our prediction-guided feature imitation adopts the whole feature maps to imitate, \textbf{without any hand-craft region selection}. And there are also No hyper-parameters to tune, which equips our method with stronger generalization. We do not rely on GT labels to partition foreground and background either.       
(2) To our best knowledge, we are the first to mimic the rank distribution of anchors. And we validate that the rank of anchors is more effective than traditional soft labels.

\section{Methods}

\subsection{Classic Knowledge Distillation}

In this section, we briefly recap the classic knowledge distillation for object detection. Firstly, for the classification branch, the cross-entropy (CE) loss is usually incorporated to align class probabilities $p_s^i$ with one-hot labels $y_i$: 
\begin{equation}
    Loss_{CE} = -\sum_{i=1}^N{y_{i}log(p_s^i)},
\end{equation}
where $i$ refers to the $i$th sample. Besides, the class probabilities of teacher models can serve as soft labels to provide the structural information between different categories~\cite{hinton}. Kullback-Leibler (KL) divergence is incorporated to measure the distance between student probabilities $p_s^{i,c}$ and teacher probabilities $p_t^{i,c}$, where $c \in$ \{1,2,...,C\} is the class number. Through minimizing their KL divergence, we can realize the prediction alignment:
\begin{equation}
    p_s^{i,c'} = \frac{exp(p_{s}^{i,c}/T)}{\sum_{j=1}^{C}{exp(p_{s}^{i,j}/T)}} ~~~~ 
    p_t^{i,c'} = \frac{exp(p_{t}^{i,c}/T)}{\sum_{j=1}^{C}{exp(p_{t}^{i,j}/T)}} 
\end{equation}
\begin{equation}
     Loss_{KL} = -\sum_{c=1}^C{p_t^{i,c'}log(\frac{p_s^{i,c'}}{p_t^{i,c'}})}.
\end{equation}

Apart from predictions, intermediate feature maps also carry rich information~\cite{fitnets}, and forcing students to imitate intermediate features from teachers can significantly accelerate the training convergence and improve student performance~\cite{TAR,deFeat}. 
Feature Pyramid Network (FPN)~\cite{FPN} aggregates features from different stages of backbones, contributing to semantically stronger features. Some works~\cite{deFeat} empirically validate imitating features from FPN is better than imitating them from backbones, thus we also adopt FPN feature imitation in this work, which can be formally described as:
\begin{equation}
    Loss_{mse} = \frac{1}{L}\sum_{l=1}^L{\frac{1}{H_{l}*W_{l}}||F_{l}^s - F_{l}^{t}||_{2}^{2}},
\label{equ:mse}
\end{equation}
where $L$ refers to the number of FPN levels, and $H_{l},W_{l}$ refer to the height and width of the $l$th level FPN features, respectively.  

\subsection{Rank Mimicking}

We apply traditional soft labels to various detectors, and observe that soft labels can only bring negligible improvements to object detection, with results shown in Tab.~\ref{tab:KDandRM}. Similar conclusions are drawn in ~\cite{mimicking}, which claims classic logits matching does not work well in object detection. We conjecture the reasons mainly come from the following two aspects: (1) Many detectors transfer the multi-class classification into multiple binary classifications~\cite{focal,FCOS,atss,gfl}, abandoning the structural relationships between different categories. (2) In one-stage detectors, the score is used not only to distinguish different categories, but also to rank boxes for NMS. Especially for the SOTA detectors (i.e. GFL~\cite{gfl,li2021generalized} and VarifocalNet~\cite{varifocal}), they employ joint scores to represent the class probability and localization quality at the same time. Therefore, distilling soft labels for object detection is sub-optimal .

\begin{figure}[t]
    \centering
    \includegraphics[width=0.42\textwidth]{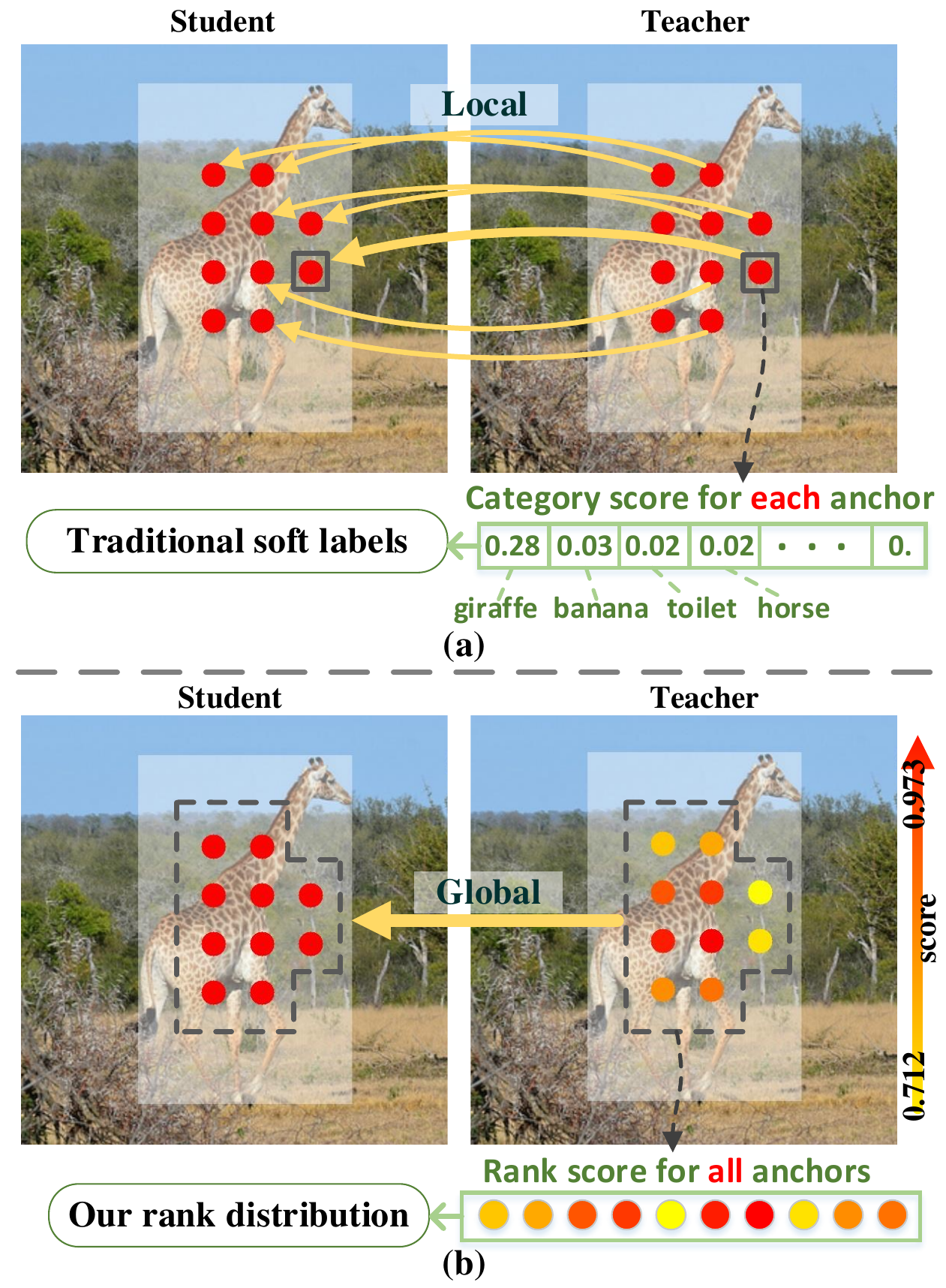}
    \caption{Comparison between traditional soft labels (a) and our rank distribution (b). 
    Yellow and red points refer to anchors.
    While traditional soft label methods locally transfer individual category score knowledge, our method can distill the global relationship, i.e., rank distribution, from teachers, leading to further performance improvement.
    }
    \label{fig:score_distribution}
\end{figure}

Through comparing the detection results between teacher and student, we find their performance gap may come from the different rank distributions of anchors. 
Some examples are shown in Fig.~\ref{fig:motivation_RM}. For easy examples, the student predicts the final detection box from the same anchor as the teacher; while for hard examples, the student lags behind the teacher. To be specific, the student produces the inferior detection box, which is regressed from a distinct anchor compared with the teacher, indicating that the rank distribution heavily affects detection performance. 

In modern object detection, for one instance, we usually assign multiple positive anchors to it. Detection models are capable of modeling the relationships between these anchors, and assigning higher scores to those high-quality anchors. 
We visualize the positive anchors along with their scores in Fig.~\ref{fig:score_distribution} (b). It is obvious that anchors located at objects are assigned with higher scores than those located at background regions.
However, the capacity of modeling anchor relations differs for the teacher and student. As Fig.~\ref{fig:motivation_RM} indicating, for easy examples, both teachers and students can model precise anchor relationships; however, for hard examples, students are weaker than teachers in modeling relations due to limited capacity.
Therefore, the rank distribution from teachers can serve as effective knowledge to transfer, enabling students to model precise relationships. 
Different from traditional soft label distillation which is conducted on each anchor locally and individually, \textbf{our rank distribution models relationships between different positive anchors for each instance in a global manner}.

We propose Rank Mimicking (RM) to transfer the rank distribution from teachers to students. Specifically, RM is performed in an instance-wise manner. For a certain instance (noted as $j$), we note its assigned $N$ positive anchors as $a_{i}^{j}$, where $i \in \{1, ..., N\}$, and their corresponding classification scores from the target category as $s_{i}^{j}$ for student and $t_{i}^{j}$ for teachers.  
To obtain the rank distribution, we apply the Softmax function to these anchor scores,  
\begin{equation}
    s_{i}^{j'} = \frac{exp(s_{i}^{j})}{\sum_{m=1}^{N}{exp(s_{m}^{j})}}~~~~
    t_{i}^{j'} =\frac{exp(t_{i}^{j})}{\sum_{m=1}^{N}{exp(t_{m}^{j})}}.
\end{equation}
Then we minimize the KL divergence between teacher and student's rank distributions:
\begin{equation}
    Loss_{RM} = -\frac{1}{M}\sum_{j=1}^{M}{ \sum_{i=1}^{N}{t_{i}^{j'}log(\frac{s_{i}^{j'}}{t_{i}^{j'}})}
    }.
\end{equation}
where M refers to the number of instances. Besides ranking candidate boxes directly by their scores, we can also rank them by the box quality (IoU with GT). 
We conduct the ablation study later in experiments to compare the two forms of rank distribution.

\subsection{Prediction-guided Feature Imitation}

Feature imitation has been proved to be a general and effective knowledge distillation method~\cite{fitnets}. However, in object detection, feature maps from multiple FPN levels are of high-dimension, and directly imitating high-dimensional information is too difficult for compact students. Therefore, selecting the most informative regions to imitate is necessary.
Previous works either imitate proposal features, or they delicately design imitation masks based on some prior, e.g., features near target objects usually carry more information than that from background regions.
However, these heuristic designs prevent feature imitation from performing in an adaptive manner, and also have the risk of falling into a local optimum.

Our intuition is that regions with large prediction differences carry important information on why students lag behind teachers. Therefore, during feature imitation, paying more attention to these regions will effectively help students catch up with teachers.
We first visualize the feature difference and prediction difference between teacher and student for vanilla feature imitation, but we find inconsistency between them, as shown in Fig.~\ref{fig:motivation_PFI}.
As the feature difference is positively correlated with the imitation supervision (as Equ.~\eqref{equ:mse}), regions with large feature differences will receive dramatic gradients during feature imitation.   
On the region inside the yellow box, prediction differences are small, which illustrates that the student can perform precise predictions as teachers by itself. Therefore, it is expected that these regions only need to receive small gradients during feature imitation and students can generate features on their own. 
However, in fact, feature differences in these regions are large, resulting in unexpected gradients which force students to generate the same features as teachers. The inconsistency between prediction differences and feature differences makes feature imitation inefficient and even impairs student training.           

Based on the above observations, we propose Prediction-guided Feature Imitation (PFI). Specifically, we incorporate the position-wise prediction difference as guidance of feature imitation, and force imitation to focus on regions with large prediction differences.   
Here, for simplicity, we only consider classification maps as prediction maps ($P_{stu} \in \mathbb{R}^{H \times W \times C}$). $P_{stu}^{c}$ refers to the predicted probability map for the class $c$. The prediction map from teachers is denoted as $P_{tea} \in \mathbb{R}^{H \times W \times C}$. The prediction differences between teacher and student are noted as:
\begin{equation}
    P_{dif} = \frac{1}{C}\sum_{c=1}^C{|P_{stu}^{c} - P_{tea}^{c}|^{2}},
\end{equation}
where $P_{dif} \in \mathbb{R}^{H \times W}$, representing the position-wise prediction difference. We also note the position-wise feature difference as: 
\begin{equation}
    F_{dif} = \frac{1}{Q}\sum_{q=1}^Q|F_{stu}^{q} - F_{tea}^{q}|^{2}, 
\end{equation}
where $F_{s}^{q}$ represents the $q$th channel of student FPN features. Then we use $P_{dif}$ to calibrate feature imitation via element-wise product:
\begin{equation}
    Loss_{PFI} = \frac{1}{L}\sum_{l=1}^L{\frac{1}{H_{l}*W_{l}}||P_{dif} \odot F_{dif}||_{2}^{2}}.
\end{equation}
\subsection{Overall loss function}

We train the student model end-to-end, and the overall loss function for distilling the student model is as follows:
\begin{equation}
    Loss = Loss_{task} + \alpha Loss_{RM} + \beta Loss_{PFI},
\end{equation}
where $Loss_{task}$ is the task loss for object detection. $\alpha$ and $\beta$ are used to balance each loss in the same scale. We empirically set $\alpha$ as 4 and $\beta$ as 1.5 by default. 

\begin{table*}[t]
    \centering
    \renewcommand{\arraystretch}{1.1}
    \resizebox{0.9\width}{!}{
    \begin{tabular}{l|l|lccccc}
    \hline
    Method & Distillation & AP & AP$_{50}$ & AP$_{75}$ & AP$_{S}$ & AP$_{M}$ & AP$_{L}$ \\ \hline
    Teacher & ResNet50-RetinaNet & 36.1 & 56.0 & 38.6 & 20.0 & 39.9 & 47.9 \\
    Student & ResNet18-RetinaNet & 31.9 & 50.4 & 33.8 & 16.6 & 34.6 & 43.8  \\ \hline
    KD: Rank Mimicking & R50-R18-RetinaNet & 33.3 (+1.4) & 51.2 & 35.6 & 17.3 & 36.3 & 44.9 \\
    KD: Prediction-guide Feature Imitation & R50-R18-RetinaNet & 34.2 (+2.3) & 53.0 & 36.5 & 17.3 & 37.6 & 46.9 \\
    KD: PFI + RM & R50-R18-RetinaNet & \textbf{34.8} (+2.9) & \textbf{53.2} & \textbf{37.2} & \textbf{17.9} & \textbf{38.1} & \textbf{47.6} \\ \hline
    \hline
    Teacher & ResNet50-FCOS & 36.6 & 55.5 & 38.7 & 20.4 & 40.6 & 47.2 \\
    Student & ResNet18-FCOS & 32.5 & 50.2 & 34.4 & 17.8 & 35.6 & 42.6 \\ \hline
    KD: Rank Mimicking & R50-R18-FCOS & 33.4 (+0.9) & 51.0 & 35.6 & 18.3 & 36.4 & 44.0 \\
    KD: Prediction-guide Feature Imitation & R50-R18-FCOS & 34.8 (+2.3) & 53.0 & 36.8 & 19.5 & 38.5 & 45.2 \\
    KD: PFI + RM & R50-R18-FCOS & \textbf{35.1} (+2.6) & \textbf{53.3} & \textbf{37.1} & \textbf{19.3} & \textbf{38.9} & \textbf{45.9} \\ \hline
    \hline
    Teacher & ResNet50-ATSS & 38.7 & 57.8 & 41.6 & 23.9 & 42.2 & 49.0 \\
    Student & ResNet18-ATSS & 34.8 & 53.1 & 37.1 & 19.3 & 37.9 & 45.61 \\ \hline
    KD: Rank Mimicking & R50-R18-ATSS & 35.4 (+0.6) & 53.3 & 37.9 & 19.3 & 38.6 & 46.1 \\
    KD: Prediction-guide Feature Imitation & R50-R18-ATSS & 36.9 (+2.1) & \textbf{55.4} & 39.4 & \textbf{21.7} & 40.2 & \textbf{48.2} \\
    KD: PFI + RM & R50-R18-ATSS & \textbf{37.1} (+2.3) & 55.3 & \textbf{39.5} & 21.0 & \textbf{40.7} & 48.1 \\ \hline
    \hline
    Teacher & ResNet50-GFL & 39.9 & 58.6 & 42.6 & 22.8 & 43.7 & 51.9 \\
    Student & ResNet18-GFL & 35.8 & 53.6 & 38.0 & 19.8 & 38.7 & 46.9 \\ \hline
    KD: Rank Mimicking & R50-R18-GFL & 36.4 (+0.6) & 53.8 & 38.7 & 19.8 & 39.3 & 48.0 \\
    KD: Prediction-guide Feature Imitation & R50-R18-GFL & 38.1 (+2.3) & 56.3 & 40.4 & 21.0 & 41.6 & 50.0 \\
    KD: PFI + RM & R50-R18-GFL & \textbf{38.3} (+2.5) & \textbf{56.3} & \textbf{40.7} & \textbf{21.8} & \textbf{41.7} & \textbf{51.0} \\ \hline
    \end{tabular}\
    }
    \vspace{-7pt}
    \caption{Results of the proposed RM and PFI on COCO dataset with different detection frameworks.}
    \label{tab:variousDet}
\end{table*}

\section{Experiments}

\subsection{Datasets and Metrics}

\noindent\textbf{COCO} is a large-scale object detection benchmark~\cite{coco}. Following the standard practice~\cite{FPN,focal}, we use the \textit{trainval135k} (115K images) for training, and \textit{minival} set (5K images) as validation. We conduct ablation studies and report main results on the \textit{minival} set. We select COCO Average Precision (AP) as the metric.

\noindent\textbf{PASCAL VOC} is also used to evaluate the robustness of our method~\cite{pascal}. 
We choose the 5k \textit{trainval} images split in VOC 2007 and 16k \textit{trainval} images split in VOC 2012 for training, and 5k \textit{test} images split in VOC 2007 for test. The performance is measured by VOC-style mAP with IoU threshold at 0.5.

\subsection{Implementation Details}

All experiments are performed in 8 GTX 1080Ti GPUs. We use SGD as optimizer with a batchsize of 2 images per GPU. For ablation studies on COCO, all models are trained for 12 epochs, knows as 1x schedule. The learning rate is set as 0.02, and divided by 10 at 8-th and 11-th epoch. Following the practices in GID~\cite{GID}, when comparing with the state-of-the-art methods (Tab.~\ref{tab:coco}), we adopt 2x schedule (24 epochs). On PASCAL VOC, models are trained for 18 epochs with an initial learning rate of 0.02, then the learning rate is divide by 10 at the 14-th and 16-th epoch. 

\begin{table}[t]
    \centering
    \renewcommand{\arraystretch}{1.1}
    \resizebox{0.44\textwidth}{!}{
    \begin{tabular}{c|l|l|l}
    \hline
     \multirow{2}{*}{KD Method} &  \multicolumn{3}{c}{mAP} \\ \cline{2-4}
     & RetinaNet & FCOS & GFL  \\ \hline
       \hline
     Teacher R50 & 36.1 & 36.6 & 39.9 \\
     Student R18 & 31.9 & 32.5 & 35.8 \\ \hline
     soft labels & 32.4 {\begin{small}{(\textbf{+0.5})}\end{small}} & 32.9 {\begin{small}{(\textbf{+0.4})}\end{small}} & 36.0 {\begin{small}{(\textbf{+0.2})}\end{small}} \\ \hline
     Rank Mimicking & \textbf{33.3} {\begin{small}{(\textbf{+1.4})}\end{small}} & \textbf{33.4} {\begin{small}{(\textbf{+0.9})}\end{small}} & \textbf{36.4} {\begin{small}{(\textbf{+0.6})}\end{small}} \\ \hline
    \end{tabular}
    }
    \vspace{-8pt}
    \caption{Comparison between traditional soft label distillation and our Rank Mimicking on COCO benchmark.}
    \label{tab:KDandRM}
\end{table}

\section{Ablation Study}

\noindent\textbf{Effectiveness of each component}. To validate the effectiveness and robustness, we implement our methods on different detectors with various backbones. 
(1) With the ResNet50~\cite{resnet} based detectors as the teacher and ResNet18 based detectors as the student, we apply Rank Mimicking (RM) and Prediction-guided Feature Imitation (PFI) to advanced detectors (RetinaNet~\cite{focal}, FCOS~\cite{FCOS}, ATSS~\cite{atss} and GFL~\cite{gfl}) step by step. The results are reported in Tab.~\ref{tab:variousDet}. Firstly, for all detectors, both RM and PFI obtain significant improvements. On RetinaNet, our RM obtains 1.4\% gains based on the baseline. And our PFI obtains at least 2.1\% gains on four detectors. Secondly, RM and PFI can complement each other. Equipped With both RM and PFI, our KD method achieves the best performance. Specifically, on RetinaNet, our method gets 2.9\% improvement, which makes up 69\% performance gap between teacher and student. 
(2) We adopt RetinaNet as the detector and validate our methods on different teacher-student pairs. Tab.~\ref{tab:t-s-pairs} indicates our methods consistently obtain significant improvements, regardless of weak (mbv2-half~\cite{mobilenets}) or strong (ResNet50) student backbones. On all teacher-student pairs, our RM gets more than 1\% improvements.  
Particularly for mobileNet-V2 (mbv2), which has different basic blocks from ResNet, our KD methods still work well.     

\begin{table}[]
    \centering
     \renewcommand{\arraystretch}{1.1}
    \resizebox{0.46\textwidth}{!}{
    \begin{tabular}{cc|c|c|c}
    \hline
    \multirow{2}{*}{RM} & \multirow{2}{*}{PFI} & R50 (36.1) & R101 (37.8) & mbv2 (29.6)  \\
     & & R18 (31.9) & R50 (36.1) & mbv2-half (25.3) \\ \hline
     \hline
     \checkmark &  & 33.3 {\begin{small}{(\textbf{+1.4})}\end{small}} & 37.1 {\begin{small}{(\textbf{+1.0})}\end{small}} & 26.7 {\begin{small}{(\textbf{+1.4})}\end{small}} \\
     & \checkmark & 34.2 {\begin{small}{(\textbf{+2.3})}\end{small}} & 38.1 {\begin{small}{(\textbf{+2.0})}\end{small}} & 27.7 {\begin{small}{(\textbf{+2.4})}\end{small}} \\
     \checkmark & \checkmark & \textbf{34.8} {\begin{small}{(\textbf{+2.9})}\end{small}} & \textbf{38.4} {\begin{small}{(\textbf{+2.3})}\end{small}} & \textbf{27.9} {\begin{small}{(\textbf{+2.6})}\end{small}} \\ \hline
    \end{tabular}}
    \vspace{-7pt}
    \caption{AP results of the proposed RM and PFI on COCO dataset with different teacher-student pairs. The 1st and 2nd rows refer to the teacher and student, respectively.
    Numbers in brackets after networks refer to their performance.}
    \label{tab:t-s-pairs}
\end{table}

\begin{table}[t]
    \centering
    \renewcommand{\arraystretch}{1.1}
    \resizebox{0.34\textwidth}{!}{
    \begin{tabular}{cc|ccc}
    \hline
    cls rank & loc rank & AP & AP$_{50}$ & AP$_{75}$  \\ \hline
    \checkmark & & 32.9 & \textbf{51.5} & 35.0 \\
    \checkmark & \checkmark & \textbf{33.3} & 51.2 & \textbf{35.6} \\\hline
    \hline
    \multicolumn{2}{c|}{S: R18-RetinaNet} & 31.9 & 50.4 & 33.8 \\ 
    \multicolumn{2}{c|}{T: R50-RetinaNet} & 36.1 & 56.0 & 38.6 \\ \hline
    \end{tabular}
    }
    \vspace{-7pt}
    \caption{Ablation study on two forms of rank distribution on COCO dataset.}
    \label{tab:ablation_RM}
\end{table}

\begin{table}[t]
    \centering
    \renewcommand{\arraystretch}{1.1}
    \resizebox{0.34\textwidth}{!}{
    \begin{tabular}{c|ccc}
    \hline
    Methods & AP & AP$_{50}$ & AP$_{75}$  \\ \hline
    whole map & 37.6 & 55.8 & 40.0  \\ 
    Positive samples & 37.2 & 55.3 & 39.5  \\
    Negative samples & 37.5 & 55.4 & 39.9 \\
    GT Mask & 37.5 & 55.5 & 40.1 \\
    PFI (ours) & \textbf{38.1} & \textbf{56.3} & \textbf{40.4} \\ \hline
    \hline
    S: R18-GFL & 35.8 & 53.6 & 38.0 \\ 
    T: R50-GFL & 39.9 & 58.6 & 42.6 \\ \hline
    \end{tabular}
    }
    \vspace{-7pt}
    \caption{Comparison between our PFI and other feature imitation methods on COCO.}
    \label{tab:imitation_mask}
\end{table}

\begin{table}[t]
    \centering
    \renewcommand{\arraystretch}{1.1}
    \resizebox{0.44\textwidth}{!}{
    \begin{tabular}{cc|cccc}
    \hline
    Methods & Distillation & AP & AP$_{S}$ & AP$_{M}$ & AP$_{L}$ \\ \hline
    Teacher & R101-RetinaNet & 38.5 & 20.6 & 42.6 & 51.6 \\
    Student & R50-RetinaNet & 36.9 & 19.9 & 40.7 & 49.0 \\ 
    \hline
    Fitnet$^{\ast}$ & R101-R50-Retina & 37.4 & 20.8 & 40.8 & 50.9 \\
    FGFI$^{\ast}$ & R101-R50-Retina & 38.6 & 21.4 & 42.5 & 51.5 \\
    GID$^{\ast}$ & R101-R50-Retina & 39.1 & \textbf{22.8} & 43.1 & 52.3 \\
    Ours & R101-R50-Retina & \textbf{39.6} & 21.4 & \textbf{44.0} & \textbf{52.5} \\ \hline
    \hline
    Teacher & R152-RetinaNet & 40.1 & 24.1 & 43.9 & 52.8 \\
    Student & R50-RetinaNet & 36.9 & 19.9 & 40.7 & 49.0 \\ 
    \hline
    FGFI$^{\dagger}$ & R152-R50-Retina & 38.9 & 21.9 & 42.5 & 52.2 \\
    DeFeat$^{\dagger}$  & R152-R50-Retina & 39.7 & \textbf{23.4} & 43.6 & 52.9 \\
    Ours  & R152-R50-Retina & \textbf{40.4} & 23.2 & \textbf{44.6} & \textbf{53.8} \\ \hline
    \end{tabular}}
    \vspace{-7pt}
    \caption{Comparison with state-of-the-art methods on COCO. $\ast$ and $\dagger$ indicate the results borrowed from GID~\cite{GID} and DeFeat~\cite{deFeat}, respectively.}
    \label{tab:coco}
\end{table}

\begin{table}[t]
    \centering
    \renewcommand{\arraystretch}{1.1}
    \resizebox{0.3\textwidth}{!}{
    \begin{tabular}{cc|cc}
    \hline
    Methods & Distillation & mAP \\ \hline
    Teacher & R101-RetinaNet & 81.9 \\
    Student & R50-RetinaNet & 81.1 \\ 
    \hline
    Fitnet$^{\ast}$ & R101-R50-Retina & 81.4 \\
    FGFI$^{\ast}$ & R101-R50-Retina & 81.5  \\
    GID$^{\ast}$ & R101-R50-Retina & 82.0 \\
    Ours & R101-R50-Retina & \textbf{82.2} \\ \hline
    \end{tabular}
    }
    \vspace{-7pt}
    \caption{Comparison with SOTA methods on VOC. $^{\ast}$ indicate that results are borrowed from GID~\cite{GID}.}
    \label{tab:voc}
\end{table}

\noindent\textbf{Comparison with traditional soft label distillation.} 
In Tab.~\ref{tab:KDandRM}, we compare two output distillation methods: traditional soft label distillation~\cite{hinton} and our rank mimicking. It is obvious that soft labels can only bring negligible improvements to detectors (around 0.2\%$\sim$0.5\% AP). And our RM can consistently surpass soft labels on three detectors, indicating the rank mimicking is a more effective distillation method for object detection.

\noindent\textbf{Comparing two types of rank distribution}. Object detection consists of two tasks: classification and regression, thus the candidate boxes can be ranked based on two indicators: classification score and predicted box quality. We compare these two types of rank distribution in Tab.~\ref{tab:ablation_RM}. When only mimicking the rank distribution of classification scores, it can bring 1.0\% AP gains to the baseline. And we further ask students to mimic the rank distribution of predicted box quality, it can obtain 33.3\% AP, surpassing the baseline by 1.4\%. These results validate both two types of rank distribution can serve as effective knowledge to distill, and mimicking both of them achieves the best performance.


\noindent\textbf{Comparison with other feature imitation masks.}
We list some common feature imitation masks in Tab~\ref{tab:imitation_mask}. We firstly imitate the whole feature maps from FPN and get a 1.8\% AP improvement. Then, we partition all training samples into positive and negative samples according to the label assignment, and we imitate features from positive samples and negative samples respectively. The results suggest imitating only positive or only negative is inferior to imitating the whole feature maps. 
We also imitate features inside the ground-truth boxes and obtains 37.5\% AP.
Compared with these hand-crafted imitation masks, our PFI surpasses them by a large margin, at least 0.5\% AP, which benefits from the adaptive imitation region selection and tight correlation between feature imitation and prediction accuracy. 

\section{Comparison with State-of-the-art Methods}

We compare our method with other state-of-the-art distillation methods on COCO~\cite{coco} and VOC~\cite{pascal} benchmarks. For a more comprehensive comparison on COCO, we adopt two teacher-student pairs, including R152-R50 (used in DeFeat~\cite{deFeat}) and R101-R50 (used in GID~\cite{GID}). We align our training settings with DeFeat and GID, then collect results from their papers. As Tab.~\ref{tab:coco} indicates, our method surpasses previous SOTAs with a large margin (at least 0.5\% AP) for two teacher-student pairs. Fitnet~\cite{fitnets}, FGFI~\cite{FGFI} and GID~\cite{GID} all select feature imitation regions based on human priors, and our method surpasses them by 2.2\%, 1.0\% and 0.5\% respectively, indicating our prediction-guided manner is superior to their hand-craft designs. DeFeat~\cite{deFeat} delicately sets loss weights to control the influence of foreground and background regions in the distillation process, and obtains 39.7\% AP. Compared with it, our method does not rely on GT labels to explicitly partition foreground and background, and does not need to fine-tune the hyper-parameters either. Finally, our method obtains 40.4\% AP, outperforming DeFeat by 0.7\%.
Comparison results on PASCAL VOC are reported in Tab.~\ref{tab:voc}. Through the distillation, ResNet50-based RetinaNet improves the baseline from 81.1\% to 82.2\%, obtaining state-of-the-art performance. 

From Tab.~\ref{tab:coco}, we can observe our method lags behind GID on AP$_{S}$. Because GID adopts RoIAlign~\cite{maskrcnn} to resize features of different sizes into the same size during the imitation. 
However, for pixel-wise distillation, small-scale objects usually receive fewer supervisions. Addressing the scale variance for pixel-wise distillation goes beyond the scope of this work, and we leave it for further research.           

\section{Conclusion}

In this paper, through studying the behaviour difference of teacher and student, we propose rank mimicking and prediction-guided feature imitation for object detection distillation. RM incorporates a novel type of knowledge, namely rank distribution, as the distillation target, which consistently outperforms traditional soft label distillation. Besides, the proposed PFI attempts to correlate feature imitation with prediction accuracy, which contributes to efficient feature imitation. Extensive experiments on COCO and VOC validate our effectiveness and generalization.

\bibliography{aaai22}

\newpage

\appendixpage

\begin{figure*}[t]
    \centering
    \includegraphics[width=0.96\textwidth]{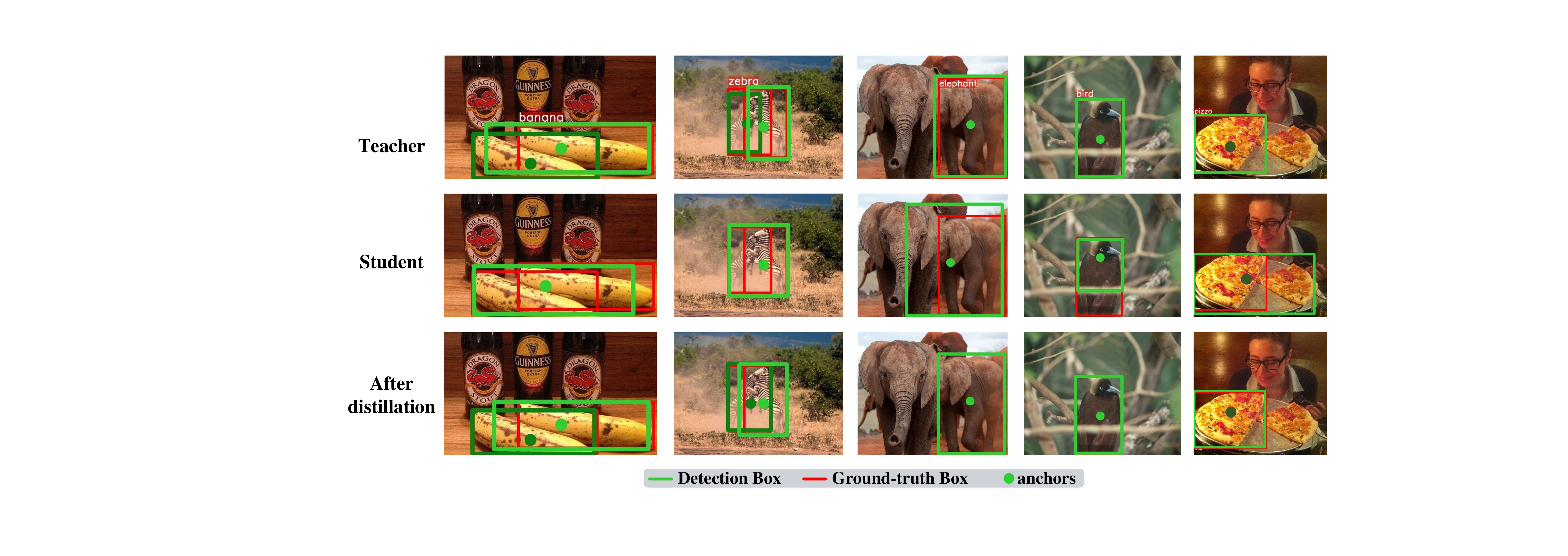}
    \caption{Qualitative results of the proposed knowledge distillation method. We visualize the ground-ground boxes (red) and detection boxes (green) along with their corresponding anchors. Teacher and Student refer to ResNet50-based and ResNet18-based RetinaNet, respectively.}
    \label{fig:Qualitative}
\end{figure*}

\section{Qualitative Results}

We show the qualitative results in Fig.~\ref{fig:Qualitative}, where the student fails to predict precise boxes for these hard examples. Compared with the teacher, the student produces final detection boxes from different anchors, which indicates the teacher and student have distinct rank distributions on these examples.  
Through the proposed knowledge distillation, rank distributions of the student successfully approach that of the teacher, consequently, the student can predict as precise detection boxes as the teacher.

\end{document}